\documentclass[conference]{IEEEtran}
\IEEEoverridecommandlockouts

\usepackage{cite}
\usepackage{amsmath,amssymb,amsfonts}
\usepackage{algorithmic}
\usepackage{graphicx}
\usepackage{textcomp}
\usepackage{xcolor}
\def\BibTeX{{\rm B\kern-.05em{\sc i\kern-.025em b}\kern-.08em
    T\kern-.1667em\lower.7ex\hbox{E}\kern-.125emX}}

\IEEEoverridecommandlockouts
\makeatletter
\def\ps@IEEEtitlepagestyle{%
\def\@oddfoot{\mycopyrightnotice}%
\def\@evenfoot{}%
}
\def\mycopyrightnotice{%
{\footnotesize \copyright~2024~IEEE. Personal use of this material is permitted. Permission from IEEE must be obtained for all other uses. \hfill} 
\gdef\mycopyrightnotice{}
}
\begin{document}

\title{Performance Evaluation of Deep Learning-Based State Estimation: A Comparative Study of KalmanNet
}

\author{\IEEEauthorblockN{Arian Mehrfard\IEEEauthorrefmark{1}\IEEEauthorrefmark{2}, Bharanidhar Duraisamy\IEEEauthorrefmark{1}, Stefan Haag\IEEEauthorrefmark{1}, Florian Geiss\IEEEauthorrefmark{1}}\\
\IEEEauthorblockA{\IEEEauthorrefmark{1} Research and Development, Mercedes-Benz AG, Germany, Email: [firstname].[lastname]@mercedes-benz.com}
\IEEEauthorblockA{\IEEEauthorrefmark{2} Department of Engineering University of Technology Nuremberg, Germany}
}

\maketitle

\begin{abstract}
Kalman Filters (KF) are fundamental to real-time state estimation applications, including  radar-based tracking systems used in modern driver assistance and safety technologies. In a linear dynamical system with Gaussian noise distributions the KF is the optimal estimator. However, real-world systems often deviate from these assumptions. This deviation combined with the success of deep learning across many disciplines has prompted the exploration of data driven approaches that leverage deep learning for filtering applications. These learned state estimators are often reported to outperform traditional model based systems. In this work, one prevalent model, KalmanNet, was selected and evaluated on automotive radar data to assess its performance under real-world conditions and compare it to an interacting multiple models (IMM) filter. The evaluation is based on raw and normalized errors as well as the state uncertainty. The results demonstrate that KalmanNet is outperformed by the IMM filter and indicate that while data-driven methods such as KalmanNet show promise, their current lack of reliability and robustness makes them unsuited for safety-critical applications.

\end{abstract}


\section{Introduction}
The advent of advanced driver assistance and safety systems (ADAS) and the ongoing development of autonomous vehicles is transforming the automotive industry. Accurate real-time perception of the environment with advanced sensor technologies is crucial for ADAS. Radar sensors are particularly important due to their robustness towards weather conditions, long detection range, and their unique ability to directly measure radial velocity. However, the data obtained from radar sensors is subject to noise and inaccuracies, thus requiring filtering to estimate its state from the noisy observations.
The Kalman Filter (KF) has, since its development by R.E. Kalman in 1960 \cite{kalman1960new}\cite{kalman1961new}, been the state of the art for state estimation under noisy observations. KF takes a probabilistic approach, predicting the future state and correcting its predictions with new measurements. Additionally, KF assumes a linear and Gaussian state space (SS) model with white Gaussian noise processes. When these conditions and assumptions are met, the KF is the optimal minimum mean square error (MMSE) estimator for dynamical systems. Even if the noise processes and state do not follow a Gaussian distribution, the KF remains the best linear MMSE estimator\cite{bar2004estimation}. However, many real-world applications are characterized by a non-linear dynamical system. Thus, extensions to the original KF were proposed, the extended Kalman filter (EKF)\cite{bar2004estimation} and the Unscented Kalman Filter (UKF)\cite{julier1997new}. The KF, EKF, and UKF are model-based (MB) algorithms and their performance relies on accurate system modeling and noise characterization as well as valid assumptions and conditions. Performance degradation occurs when the SS model diverges from the underlying dynamics and statistics\cite{revach2022kalmannet}.

Deep neural networks (DNNs) and their advancements in the past years have considerably impacted numerous fields, such as computer vision\cite{voulodimos2018deep} and natural language processing\cite{young2018recent}. The ability of DNNs to capture complex structures from data replaces the need for accurate system modeling and feature design\cite{lecun2015deep}. Consequently, it is logical to approach state estimation with data-driven (DD) DNN models. Recurrent neural networks (RNNs)\cite{chung2014empirical, Sherstinsky_2020} and transformers \cite{vaswani2017attention}, have proven reliable and effective at modeling temporal dependencies in sequential data\cite{cheng2023machine}. However, these models require a substantial number of trainable parameters and training samples and they lack the interpretability of MB methods\cite{pmlr-v70-zaheer17a}. These limitations of purely DD deep learning methods have inspired hybrid approaches that combine the interpretability of MB algorithms with the adaptability and learning capacity of neural networks\cite{shlezinger2022model, revach2022kalmannet}.Revach et al. \cite{revach2022kalmannet} propose KalmanNet, a MB and DD hybrid architecture. KalmanNet integrates RNN's into the KF equations, resulting in an interpretable state estimation model that can learn complex dynamics and noise processes. The authors benchmark their hybrid filter against the performance of an EKF on two simulations and one real-world scenario. In the real-word evaluation, odometry data of a small vehicle from the NCLT dataset \cite{ncarlevaris-2015a} is used to estimate the ego-position, the estimate is compared to the ground truth GPS position. The evaluation on this real world data is only briefly described and does not contain enough details to be a suitable benchmark. KalmanNet has been extended by Han et al. \cite{han2023multi} who propose a Multi-Model KalmanNet that utilizes multiple KalmanNets with different motion models. The estimates of these KalmanNets are mixed using weights that are predicted by a multi-layer perceptron that learns to evaluate the similarity between the predicted and observed target motion. In this work we will consider the original contribution by Revach et al. \cite{revach2022kalmannet} to allow a focused evaluation of KalmanNet.

\section{Problem Statement}
In the current literature, proposed DD state estimation models are frequently insufficiently evaluated on real-world data. KalmanNet\cite{revach2022kalmannet} by Revach et al. is a state of the art hybrid DD and MB filter with significance and high visibility in the community, therefore we select it for a detailed evaluation and performance analysis.

In this work we evaluate and compare the performance of a KalmanNet model, trained on real-world automotive data, against the interacting multiple model (IMM) algorithm. The KalmanNet architecture is employed and trained on radar-data from the RadarScenes \cite{radar_scenes_dataset} dataset, and its state estimation is systematically compared to that of an IMM algorithm.

\section{Relevant Work}
This section provides background information on the linear KF, which is the foundation of the IMM algorithm and the KalmanNet architecture. The KF is an algorithm that estimates the unknown state ($x$) of a target under noisy observations ($z$). 
The targets behaviour can be characterized by a discrete-time linear dynamical system with a motion and observation model:
\begin{flalign}
&\hat{x}_{k+1}=F_{k\Delta t}\hat{x}_k + v_k \label{eq:kf:mot}\\
&z_k = H_k\hat{x}_k + w_k \label{eq:kf:obs}
\end{flalign}

In \eqref{eq:kf:mot} $\hat{x}_{k+1} \in \mathcal{R}^m$ is the predicted state vector at step $k+1$, which is propagated from the state at step $k$ using the linear state transition function $F$, disturbed by the white Gaussian process noise $v_k \sim \mathcal{N} (0, \mathbf{Q})$ with covariance $Q$. At step $k$, the observation $z_k \in \mathcal{R}^n$ in \eqref{eq:kf:obs} is given by the linear measurement function $H_k$ and corrupted by white Gaussian noise $w_k \sim \mathcal{N} (0, \mathbf{R})$ with covariance $R$.    
The KF\cite{bar2004estimation} is an algorithm that estimates the state of a dynamical system, given that the transition and observation function are linear and the noise follows a Gaussian distribution. Furthermore, the initial state $x_0$ is modeled as a Gaussian distributed random variable, described by the first raw and second central statistical moments, a property that is preserved through the linearity of $F$ and $H$. As a recursive state estimator, the KF operates in a prediction-update loop, where the prediction propagates the state $\hat{x}_{k|k}$ and its covariance $P_{k|k}$ to $\hat{x}_{k+1|k}$ and $P_{k+1|k}$, the latter two being $\hat{x}$ and $P$ at time $k+1$ given information up to time $k$:
\begin{flalign}
&\hat{x}_{k+1|k}=F_{k}\hat{x}_{k|k} \label{eq:kf:pred:x}\\
&P_{k+1|k} = F_k P_{k|k}F_k' + Q_k \label{eq:kf:pred:p}
\end{flalign}

Analogously, the predicted measurement $\hat{z}_{k+1|k}$ and the predicted measurement covariance $S_{k+1}$ are computed as:
\begin{flalign}
&\hat{z}_{k+1|k}=H_{k+1}\hat{x}_{k+1|k} \label{eq:kf:pred:z} \\
&S_{k+1} = H_{k+1} P_{k+1|k}H_{k+1}' + R_k \label{eq:kf:pred:z:cov}
\end{flalign}

Using \eqref{eq:kf:pred:p} and \eqref{eq:kf:pred:z:cov} the Kalman Gain (KG) $W$ is given by:

\begin{equation}
    W_{k+1} = P_{k+1|k} H_{k+1}' S_{k+1}^{-1} \label{eq:kf:kg}
\end{equation}

During the update phase of step $k+1$ the measurement $z_{k+1}$ is obtained and combined with \eqref{eq:kf:pred:z}, forming the innovation or measurement residual $\nu$:
\begin{equation}
    \nu _{k+1} = z_{k+1} - \hat{z}_{k+1|k} \label{eq:kf:innov}
\end{equation}
Combining the above equations, the posterior first and second statistical moment of the state can be computed:

\begin{flalign}
    &\hat{x}_{k+1|k+1} = \hat{x}_{k+1| k} + W_{k+1}\nu _{k+1} \label{eq:kf:x:up} \\
    &P_{k+1|k+1} = P_{k+1|k} - W_{k+1} S_{k+1} W_{k+1}' \label{eq:kf:p:up}
\end{flalign}

Under the assumptions of Gaussian distributions and linearity, the KF is the optimal minimum mean square error state estimator\cite{bar2004estimation}. However, the KF's performance is dependent on the accuracy of the chosen dynamical model. The most commonly used models are the constant velocity (CV), constant acceleration (CA), and constant turn-rate (CT) models\cite{bar2004estimation}.
To address this limitation, the IMM algorithm incorporates multiple motion models, enabling more robust state estimation.

\subsection{Interacting Multiple Model Algorithm}
Operating under the assumption that the target can change its behaviour at any timestep, the IMM algorithm \cite{bar2004estimation} combines multiple dynamical models to represent varying target behaviour. Each of the dynamical models is used as a motion model for a separate KF. The IMM algorithm adapts to a change in target behaviour by switching to a different model that more accurately represents the target behaviour. This change in target behaviour is modeled as a Markov process. A cycle of an IMM model at time-step $k$ with $r$ models starts by initializing the filters with mixed state estimates from step $k-1$. Here the subscript ${k-1}$ is short for ${k-1|k-1}$:
\begin{flalign}
    &\hat{x}^{0j}_{k-1|k-1} = \sum^{r}_{i=1} \hat{x}^i_{k-1|k-1} \mu _{k-1|k-1}^{i|j} \label{eq:imm:init:mix} \\
    &P^{0j}_{k-1} = \sum^r_{i=1} \mu_{k-1}^{i|j}\{ P^i_{k-1} + (\hat{x}^i_{k-1} - \hat{x}^{0j}_{k-1})(\hat{x}^i_{k-1} - \hat{x}^{0j}_{k-1})' \}
\end{flalign}
For model $M^j$, the initial state estimate $\hat{x}^{0j}_{k-1}$ and its covariance $P^{0j}_{k-1}$, are computed as a sum of the previous estimates weighted by the mixing probability $\mu _{k-1|k-1}^{i|j}$, which is the conditional probability that model $M^i$ was used at $k-1$ given that $M^j$ will be used at $k$, conditioned on $Z_{k-1}$ with $i,j=1,...,r$:  
\begin{equation}
    \mu _{k-1}^{i|j} = \frac{1}{\Bar{c}_j}p_{ij} \mu _{k-1}^i \label{eq:imm:init:u}
\end{equation}
The Markov state transition probability $p_{ij}$, set a priori, defines the probability of switching from model $M^i$ to model $M^j$. The model probability $\mu ^i_{k-1}$ \eqref{eq:imm:update:u} is the probability that $M^i$ was the correct model at $k-1$ and can be written as $P \{ M^i_{k-1} | Z_{k-1} \}$. This is normalized by $\Bar{c}_j$
 \begin{equation}
     \Bar{c}_j = \sum ^r_{i=1} p_{ij} \mu_{k-1}^i \label{eq:imm:u:norm}
 \end{equation}
to ensure a total probability of $1$.

Next all filters run through the prediction and update cycle using the mixed state \eqref{eq:imm:init:mix} and $z_k$. Using the mixed state estimates and covariance matrix from each model, the likelihood of model $M^j$ for input $z_k$ is computed as
\begin{equation}
    \Lambda _k^i =  \mathcal{N}[z_k; \hat{z}^j[k|k-1; \hat{x}^{0j}_{k-1}], S^j[k, P^{0j}_{k-1}]] \label{eq:imm:likelihood}
\end{equation}

The model probabilities $\mu _k^j$ are updated as:
\begin{equation}
    \mu _k^j = \frac{1}{c} \Lambda _k^j \Bar{c}_j \label{eq:imm:update:u}
\end{equation}
which is normalized by 
\begin{equation}
    c = \sum _{j=1}^r \Lambda_k^j \Bar{c}_j \label{eq:imm:update:u:norm}
\end{equation}

Concluding the IMM cycle, the updated state vector and state covariance matrix are computed by combining the estimates from the individual models weighted by the model probabilities:

\begin{flalign}
    &\hat{x}_{k|k} = \sum ^r_{j=1} \hat{x}^j \mu_k^j \label{eq:imm:out:x} \\
    &P_{k|k} = \sum _{j=1}^r \mu_k^j \{ P^j_{k} + [\hat{x}_k^j - \hat{x}_k] [\hat{x}_k^j - \hat{x}_k]'\} \label{eq:imm:out:p}
\end{flalign}

\subsection{KalmanNet} \label{sec:rela:knet}
KalmanNet\cite{revach2022kalmannet} is a hybrid architecture combining classical model based algorithms with learned, data-driven components. According to the authors, it can be used for real-time state estimation in non-linear dynamical systems with only partial domain knowledge. Specifically, KalmanNet is supposed to work with approximated dynamical models and without knowledge of the process noise $v_k$ and measurement noise $w_k$. To this end, KalmanNet uses a recurrent neural network (RNN) architecture that learns to predict the Kalman Gain (KG) from data. The RNN is integrated into the KF algorithm, meaning that KalmanNet also operates in a predict and update cycle. During the prediction \eqref{eq:kf:pred:x} is used to predict the prior for the current timestep $\hat{x}_{k+1|k}$ and \eqref{eq:kf:pred:z} is used to obtain the predicted measurement $\hat{z}_{k
1|k}$. However, unlike the KF, KalmanNet does not predict the second statistical moment of the state. During the update phase, KalmanNet computes the state estimate $\hat{x}_{k+1|k+1}$ as described by \eqref{eq:kf:x:up}. Instead of explicitly computing the KG from the state and innovation covariance matrices as in \eqref{eq:kf:kg}, KalmanNet predicts the KG $W_{k+1}$ using an RNN. Thus KalmanNet operates without explicit knowledge of the state covariance matrix $P$ or the innovation covariance matrix $S$. 

The KG computation \eqref{eq:kf:kg} is based on the state and innovation covariance, the input features for the RNN module are designed to capture these statistical informations:
\begin{itemize}\label{input_features}
    \item[\textit{F1}] The \textit{observation difference} $\Delta \Tilde{z}_k = z_k - z_{k-1}$
    \item[\textit{F2}] The \textit{innovation difference} $\Delta z_k = z_k - \hat{z}_{k|k-1}$
    \item[\textit{F3}] The \textit{forward evolution difference} $\Delta \Tilde{x}_k = \hat{x}_{k|k} - \hat{x}_{k-1|k-1}$
    \item[\textit{F4}] The \textit{forward update difference} $\Delta \hat{x}_k = \hat{x}_{k|k} - \hat{x}_{k|k-1}$
\end{itemize}
Feature \textit{F3} is the difference between two consecutive posterior state estimates and \textit{F4} is the difference between the posterior and prior state estimate of the same timestep. Consequently, the available features at time $k$ are $\Delta \Tilde{x}_{k-1}$ and $\Delta \hat{x}_{k-1}$. Similarly, \textit{F1} and \textit{F2} are the difference between two consecutive observations and the difference between the observation and the predicted observation, respectively. The features \textit{F1} and \textit{F3} contain information on the change in state estimates and observations over time, while \textit{F2} and \textit{F4} encapsulate the measurement and estimate residuals. 

To match the recursive properties of the KF, the authors choose RNN's as an architecture, arguing that the internal memory of an RNN would enable it to track the covariances\cite{revach2022kalmannet}. The RNN module contains three gated recurrent units (GRU) \cite{chung2014empirical}. These three GRUs represent three covariances, the process noise covariance $\mathbf{Q}$, the state covariance $\mathbf{P}_{k|k-1}$, and the innovation covariance $\mathbf{S}_k$. The first and second GRU have a hidden state dimension of $m^2$ and the third GRU of $n^2$. The three GRU's are complemented with dedicated input and output fully connected layers and are interconnected such that the output of the $\mathbf{Q}$ GRU is used as an input to the $\mathbf{P}_{k|k-1}$ GRU, which is in turn used as input for the $\mathbf{S}_k$ GRU, with the latter two being used to predict the KG. 

During training, KalmanNet is supervised on the posterior state estimate $\hat{x}_k$ instead of the RNN's predicted KG. The loss function is a least error loss:
\begin{equation}
    \mathcal{L} = \| x_k - \hat{x}_{k|k}\|^2
\end{equation}

More details on the architecture, training, and the original code can be found in \cite{revach2022kalmannet}.

\section{Training Setup \& Data}
The main contribution of this work is a comprehensive comparative analysis of the KalmanNet system in an open-loop evaluation using real-world automotive radar data from the RadarScenes \cite{radar_scenes_dataset} dataset. The authors' code was adapted to suit the use case, with modifications to the loss function, to account for the domain specific radial distribution of radar data. We split our data into training, validation, and test sets to train and evaluate KalmanNet 

\subsection{Dataset}
RadarScenes \cite{radar_scenes_dataset} contains labeled radar- and corresponding odometry data. Details on the sensor setup can be found in \cite{radar_scenes_dataset, haag2020oafuser}. In addition to the measured radial distance, azimuth, range rate, and the radar cross section, the data also contains ego-motion compensated radial velocity, Cartesian $x$ and $y$ coordinates in the ego vehicle (EV) coordinate system, as well as a global coordinate system. The data was recorded from four radar sensors mounted on the EV. For each sensor the dataset contains a transformation matrix describing the sensor mounting position. Additionally, each point has a \textit{label id}, which is the semantic class id of the object from which the measurement originated, and a \textit{track id} which uniquely associates the point to an object in the scene. To provide an isolated evaluation of the state estimation performance, clustering, data association and track management are not considered in the evaluation. To this end, the ground truth is used to create single target input sequences for each motorized vehicle with at least four wheels. Each single target input sequence consists of $K$ timesteps, where every timestep $k \in K$ contains a cluster of radar points that belong to the same target. This work uses the Cartesian $x$ and $y$ coordinates from the dataset, the doppler velocities were pre-processed and converted to a Cartesian velocity vector.

Our training dataset consists of a total of $3057493$ radar measurements distributed on $5003$ sequences. Our validation dataset consists of a total of $1306424$ radar points distributed over $2074$ sequences. Each sequence has a length of $K=100$.

\subsection{KalmanNet Initialization}
In this work, KalmanNet is used for a point-object birds eye view tracking. The state-space vector is $\mathbf{x} = (x, y, \dot x, \dot y, \ddot x, \ddot y)^T$, only the position and velocity are observed. Thus, to approximate the motion of the target, the state transition matrix of the constant acceleration dynamical model \cite{bar2004estimation} is used as the $\mathbf{F}$ matrix and a $4\times4$ identity matrix as the observation matrix $\mathbf{H}$.

The initial state $x_0$ is set to the first measurement.  Similarly to the original KalmanNet work\cite{revach2022kalmannet}, the hidden states of the three GRU's are initialized with zero matrices. The $\Delta t$ for the prediction is based on the timestamps of the current and previous measurement.  

\subsection{Training}
During the training, at each timestep $k$, a collection of $b$ points is passed as input to the filter. As described in \ref{sec:rela:knet}, the state will be predicted from $k-1$ to $k$. The update function is invoked sequentially for each of the $b$ input points, with new input features and RNN KG predictions. The $b$ radar measurements are chosen as follows: Should the cluster of detected radar points on the observed vehicle (OV) be less than three, all detected points and their mean are given as input to the filter. Should more than three points be detected, the points are chosen based on the compensated radial Doppler velocity. The point with the minimum, maximum, and median velocity, as well as the mean of these three points is given as input to KalmanNet. The posterior estimate for the timestep $k$ is the updated state after the $b$ updates. 

After the filter iterates through the input sequence the loss is evaluated. A mean square error loss function is used to evaluate the loss on the position. To adapt the loss function to the radar domain, the loss is scaled with the distance to the OV:
\begin{equation}
    \mathcal{L} = \frac{1}{K} \sum_{k=1}^K w(d_i) \cdot \| x_k - \hat{x}_{k|k}\|^2 \label{eq:myknet:loss}
\end{equation}
with a linearly interpolated weight
\begin{equation}
w(d) = 
\begin{cases} 
w_{\max} & \text{if } d \leq d_{\min} \\
w_{\max} + \alpha (d - d_{\min}) & \text{if } d_{\min} < d < d_{\max} \\
w_{\min} & \text{if } d \geq d_{\max} 
\end{cases}
\end{equation}
and 
\begin{equation}
\alpha = \frac{w_{\min} - w_{\max}}{d_{\max} - d_{\min}}
\end{equation}
A minimum and maximum weight is set as $w_{min}=0.4$, $w_{max}=1.0$ and interpolated for the distances between $d_{min}=20$ and $d_{max}=120$. KalmanNet \cite{revach2022kalmannet} is trained using the Adam optimizer\cite{kingma2014adam}.

\section{Results}

During the evaluation the mean absolute error (MAE) and root mean square error (RMSE) are computed for a position vector $\mathbf{x} = (x, y)$ and its estimation errors $\Tilde{x} = x - \hat{x}$ and $\Tilde{y} = y - \hat{y}$:
\begin{align}
\text{MAE}(\mathbf{\Tilde{x}}) &= \frac{1}{K} \sum_{k=1}^{K} (\left| \Tilde{x}_k \right| + \left|\Tilde{y}_k \right|)\\
\text{RMSE}(\mathbf{\Tilde{x}}) &= \sqrt{\frac{1}{K} \sum_{k=1}^{K} \left(\Tilde{x}^2 + \Tilde{y}^2 \right)}
\end{align}
Furthermore, the filters consistency is evaluated through the normalized estimation error squared (NEES) $\epsilon$ and normalized innovation squared (NIS) $\epsilon_{v,k}$, which are computed as:
\begin{align}
    &\epsilon_{k} = \Tilde{x}_{k}' P_{k}^{-1} \Tilde{x}_{k} \label{eq:nees} \\
    &\epsilon_{v,k} = \nu_k'S_k^{-1}\nu_k \label{eq:nis}
\end{align}
A confidence interval of $95\%$ is defined for filter consistency.

The trained KalmanNet model is evaluated on the RadarScenes\cite{radar_scenes_dataset} dataset and the Position Error is presented in Tab. \ref{tab:error:rs}. The position has a \textit{RMSE} of $1.13$, the \textit{MAE} is $0.66$ and $0.65$ for the $x$ and $y$ components with a standard deviation of $0.83$ and $0.81$, respectively. These results on the large RadarScenes dataset are complemented with a detailed evaluation on two out-of distribution sequences. In the first sequence the OV's trajectory follows the form of an $8$, this is called the $8$-drive scenario. The $8$-drive scenario is a common stress test for automotive trackers as the OV exhibits, challenging, highly non-linear motion. The second test sequence represents a common driving scenario, the EV follows the OV on a road and it is called the follow-drive scenario. These two scenarios are recorded with the same sensor-setup as the RadarScenes data, with an additional high accuracy ground truth system that provides precise position, velocity, and acceleration data, allowing for a more comprehensive evaluation. Further, an untuned IMM filter is used to track the OV on these two scenes to provide a reference for KalmanNet. KalmanNet was tested using the same procedure as during training, the radar points were clustered based on the ground-truth and passed to KalmanNet as a collection of $b$ radar points.

\begin{figure}[htbp]
\centerline{\includegraphics[width=\linewidth]{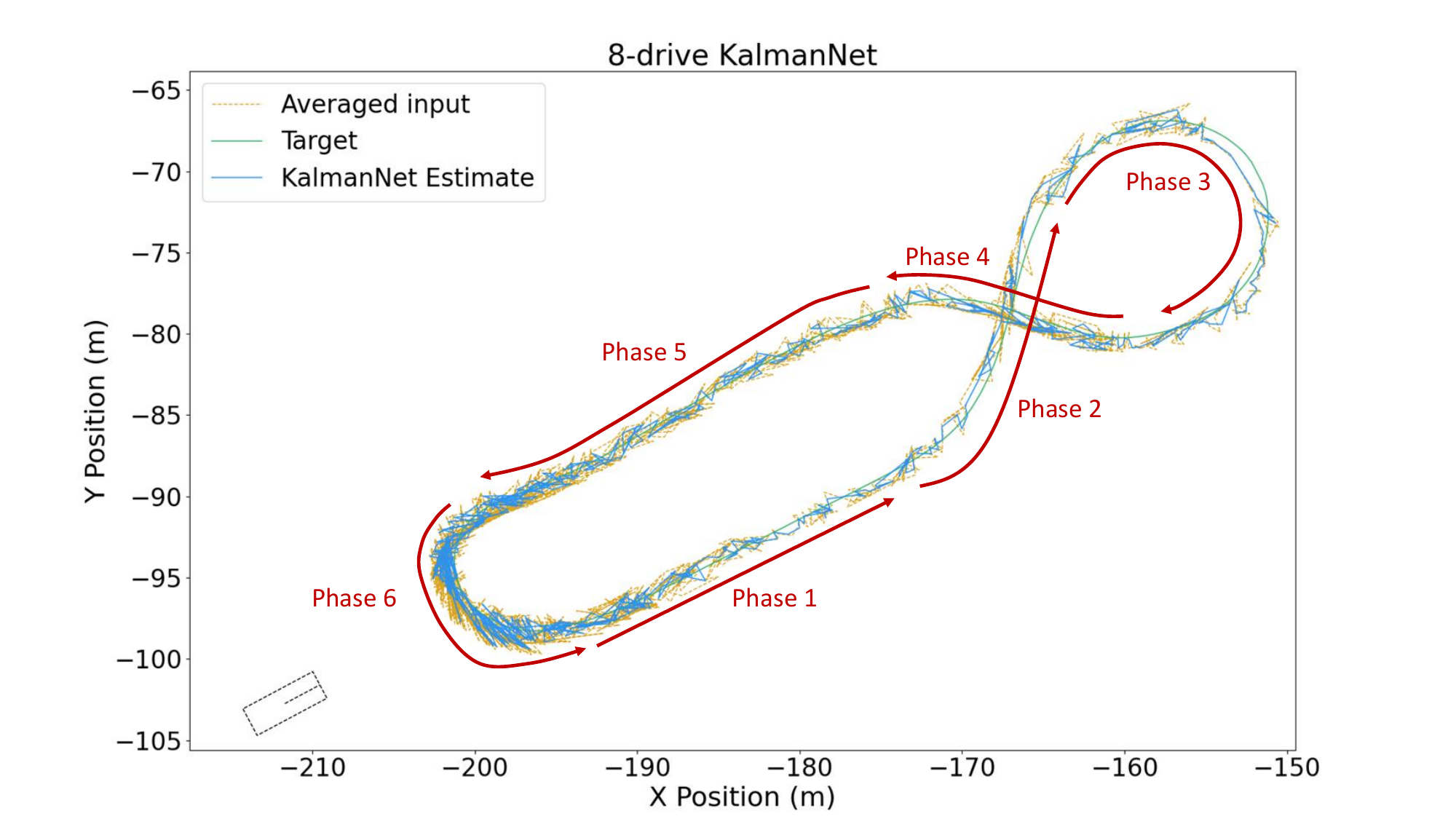}}
\caption{Estimated position of OV from KalmanNet. Input was averaged over clusters at each $k$.}
\label{fig:knet:8drive}
\end{figure}

Fig. \ref{fig:knet:8drive} visualizes the input, ground truth, and KalmanNet estimation of one $8$-drive. The red line is computed as the mean of the cluster for each time step. The green line shows the ground truth, and the blue line the estimated trajectory.

\begin{figure}[htbp]
\centerline{\includegraphics[clip, trim=2cm 0cm 2cm 2cm,width=\linewidth]{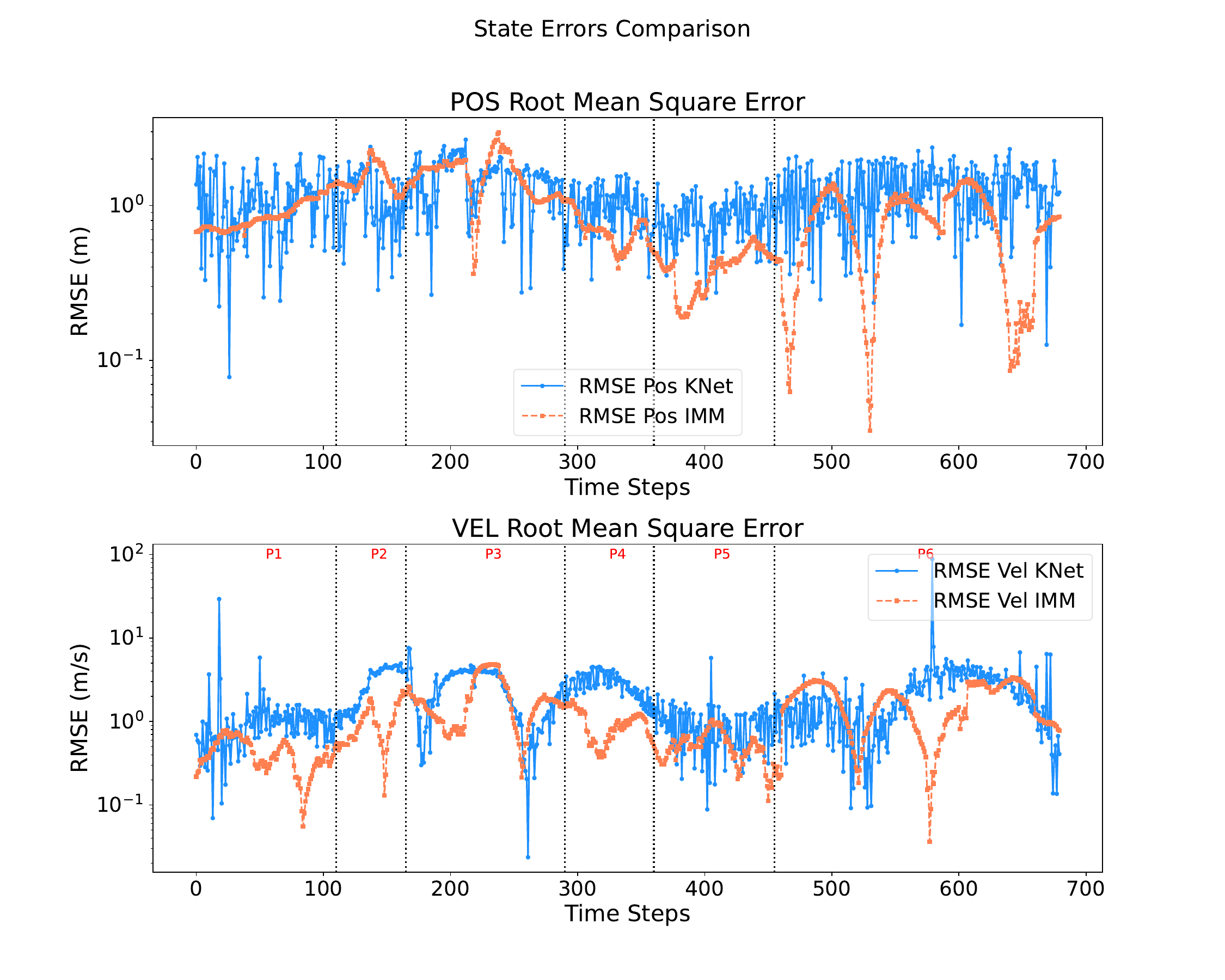}}
\caption{RMSE comparison between KalmanNet and IMM on the $8$-drive scenario. The plots shows the position (top) and velocity (bottom) errors.}
\label{fig:comp:state:8drive}
\end{figure}

\begin{figure}[htbp]
\centerline{\includegraphics[clip, trim=2cm 0cm 2cm 2cm,width=\linewidth]{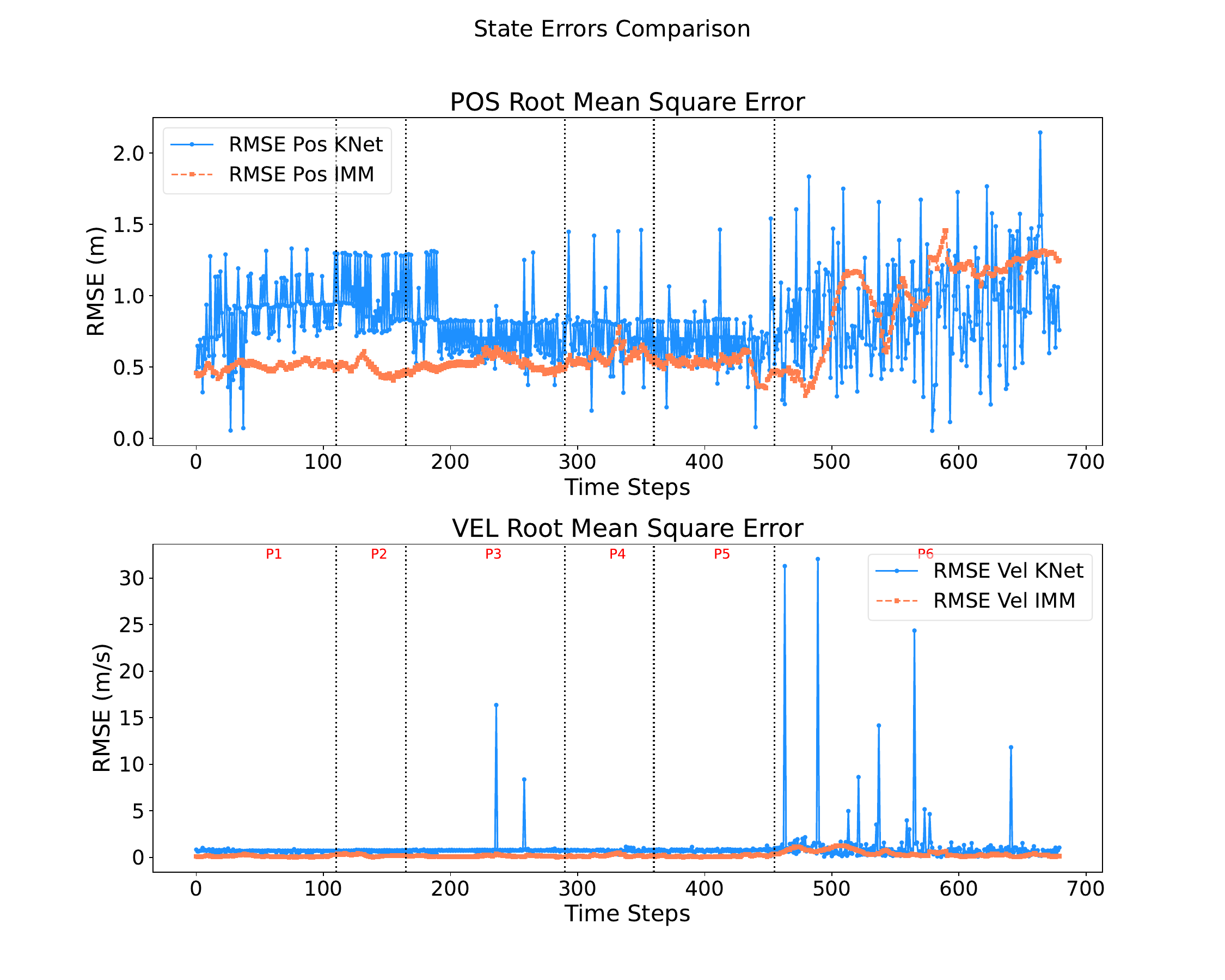}}
\caption{RMSE comparison between KalmanNet and IMM on the follow-drive scenario. The plots shows the position (top) and velocity (bottom) errors.}
\label{fig:comp:state:foll_drive}
\end{figure}

The \textit{RMSE} in position and velocity is compared between KalmanNet and IMM in Fig. \ref{fig:comp:state:8drive} for the $8$-drive, and in Fig. \ref{fig:comp:state:foll_drive} for the follow-drive scenario. The red line depicts the error of the KalmanNet algorithm and the blue line that of the IMM method. KalmanNet generally exhibits higher and more volatile errors than the IMM filter. 

\begin{figure}[htbp]
\centerline{\includegraphics[clip, trim=0cm 29.4cm 0cm 17.8cm,width=\linewidth]{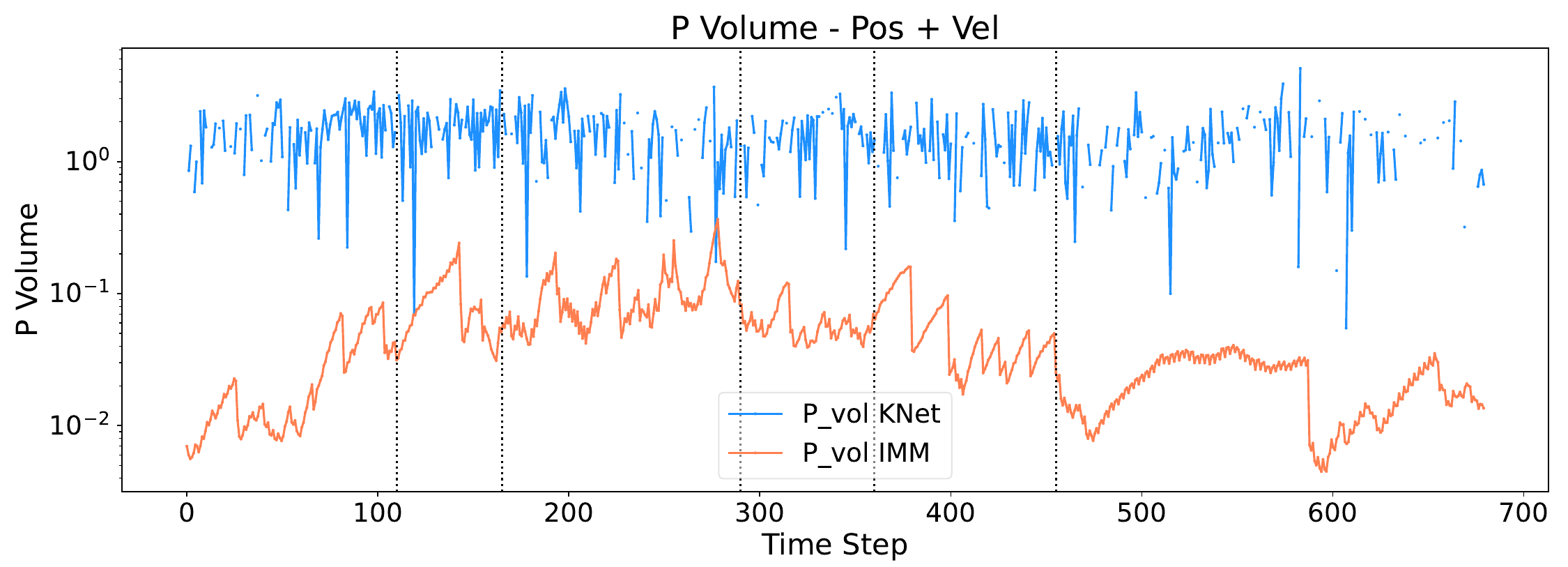}}
\caption{Comparison of the state covariance volume between KalmanNet and IMM on the $8$-drive scenario. The volume is computed for the position and velocity variance and covariance components.}
\label{fig:comp:pvol:8drive}
\end{figure}

To evaluate the state uncertainty the position and velocity components of the state uncertainty matrix are taken into consideration. Fig. \ref{fig:comp:pvol:8drive} provides a comparison of state uncertainty between KalmanNet and the IMM filter during the $8$-drive scenario, the IMM plot contains the predicted and updated covariance, whereas the KalmanNet plot only shows the updated state covariance. KalmanNet exhibits a higher state uncertainty with stronger deviations compared to the IMM state uncertainty. Both plots use a logarithmic scale.   

\begin{figure}[htbp]
\centerline{\includegraphics[clip, trim=2cm 1cm 3.5cm 2cm,width=\linewidth]{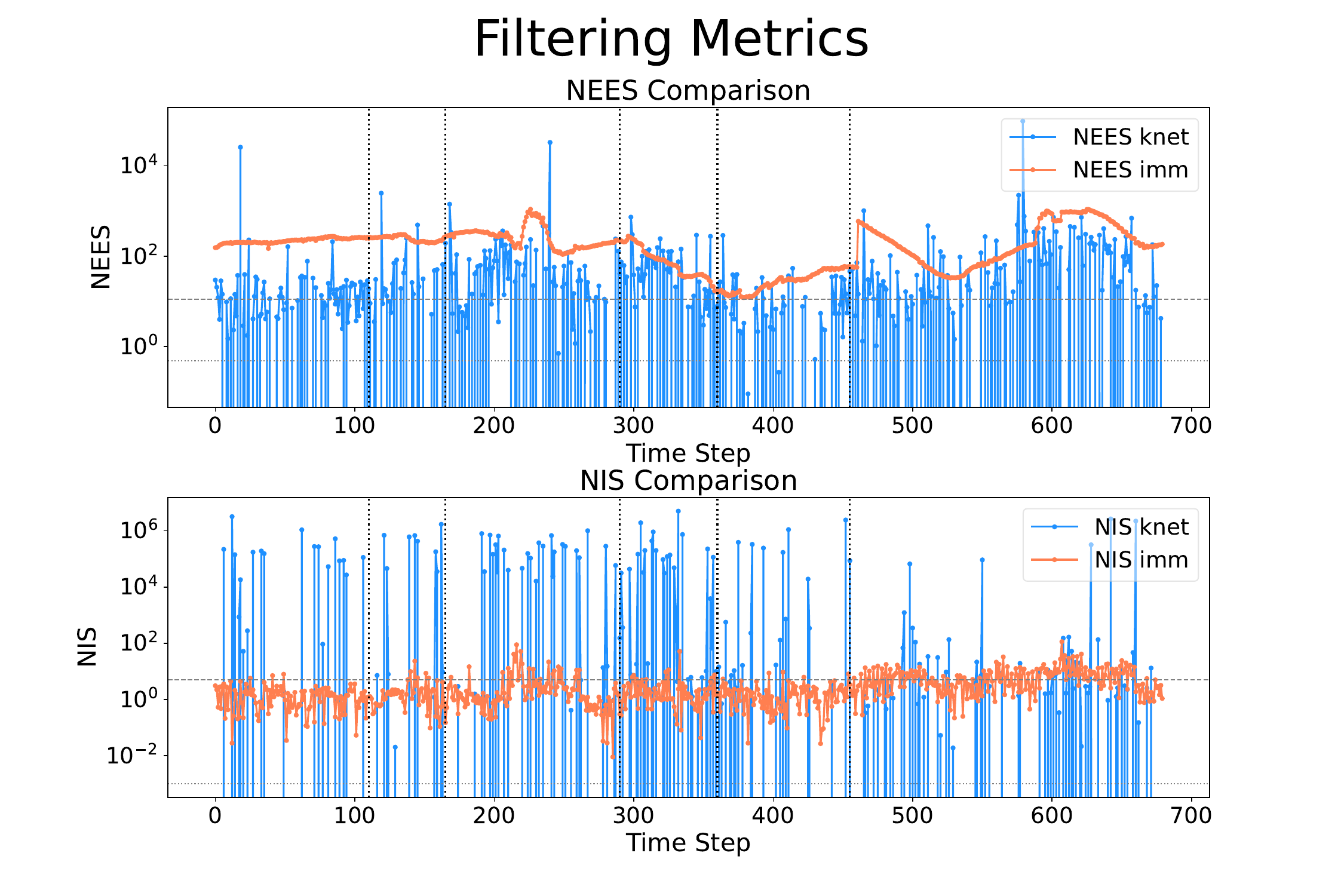}}
\caption{NEES and NIS comparison between KalmanNet and IMM on the $8$-drive scenario. Dotted lines represent the upper and lower limits of the chi-squared distributions.}
\label{fig:comp:norm:8drive}
\end{figure}

A comparison of the NEES $\nu_k$ and NIS $\nu_{v,k}$  errors for the $8$-drive scenario is shown in Fig. \ref{fig:comp:norm:8drive}. Analogously to the state uncertainty, the NEES is evaluated for the position and velocity components of the state.
The upper and lower critical values mark the $95\%$ acceptance interval of the consistency test. For the $8$-drive scenario the IMM NEES comes close to the upper bound, but remains only briefly inside the interval. The KalmanNet NEES has more points inside the interval, but it displays highly fluctuating behaviour. The IMM NIS stays inside the confidence interval for the majority of the scenario, whereas the KalmanNet NIS exhibits highly volatile behaviour. Few KalmanNet NIS errors are inside the consistency confidence interval, most NIS values are substantially above the upper critical bound.

A state-component wise comparison of the \textit{RMSE}, \textit{MAE}, and its standard deviation between KalmanNet and the IMM filter is given in Tab. \ref{tab:error:fig8} for the $8$-drive scenario and in Tab. \ref{tab:error:follow} for the follow-drive scenario. For almost every metric and state component, IMM has a lower error than KalmanNet, with a more substantial performance gap in Tab. \ref{tab:error:fig8} for the $8$-drive scenario, compared to the follow-drive in Tab. \ref{tab:error:follow}. As expected, both methods have a lower error during the follow-drive scenario than the $8$-drive.

\begin{table}[htbp]
\caption{Evaluation of KalmanNet on the RadarScenes dataset.}
\begin{center}
\begin{tabular}{c|cccccc}
\multicolumn{1}{l|}{} & \multicolumn{2}{c}{\textbf{Position}} \\
\textbf{KalmanNet}    & $\mathbf{x}$      & $\mathbf{y}$      \\ \hline
\textbf{RMSE}         & \multicolumn{2}{c}{$1.13$}            \\
\textbf{MAE}          & $0.66$            & $0.65$            \\
$\mathbf{\sigma}$     & $0.83$            & $0.81$                                     
\end{tabular}
\label{tab:error:rs}
\end{center}
\end{table}

\begin{table}[htbp]
\caption{$8$-drive scenario: Comparison of RMSE and standard deviation between KalmanNet and IMM for each state component.}
\begin{center}
\begin{tabular}{ccccccc}
\multicolumn{1}{l|}{}                   & \multicolumn{2}{c|}{\textbf{Position}}                   & \multicolumn{2}{c|}{\textbf{Velocity}}                        & \multicolumn{2}{c}{\textbf{Acceleration}}   \\
\multicolumn{1}{c|}{\textbf{KalmanNet}} & $\mathbf{x}$         & \multicolumn{1}{c|}{$\mathbf{y}$} & $\mathbf{\dot x}$    & \multicolumn{1}{c|}{$\mathbf{\dot y}$} & $\mathbf{\ddot x}$   & $\mathbf{\ddot y}$   \\ \hline
\multicolumn{1}{c|}{\textbf{RMSE}}      & \multicolumn{2}{c}{$1.23$}                               & \multicolumn{2}{c}{$2.98$}                                    & \multicolumn{2}{c}{$17.96$}                 \\
\multicolumn{1}{c|}{\textbf{MAE}}       & $0.84$               & $0.61$                            & $0.95$               & $1.37$                                 & $9.90$               & $3.98$               \\
\multicolumn{1}{c|}{$\mathbf{\sigma}$}  & $0.71$               & $\mathbf{0.62}$                   & $2.31$               & $1.86$                                 & $12.77$              & $8.37$               \\ \hline
\multicolumn{1}{l}{}                    & \multicolumn{1}{l}{} & \multicolumn{1}{l}{}              & \multicolumn{1}{l}{} & \multicolumn{1}{l}{}                   & \multicolumn{1}{l}{} & \multicolumn{1}{l}{} \\
\multicolumn{1}{c|}{\textbf{IMM}}       & $\mathbf{x}$         & \multicolumn{1}{c|}{$\mathbf{y}$} & $\mathbf{\dot x}$    & \multicolumn{1}{c|}{$\mathbf{\dot y}$} & $\mathbf{\ddot x}$   & $\mathbf{\ddot y}$   \\ \hline
\multicolumn{1}{c|}{\textbf{RMSE}}      & \multicolumn{2}{c}{$\mathbf{1.08}$}                      & \multicolumn{2}{c}{$\mathbf{1.28}$}                           & \multicolumn{2}{c}{$\mathbf{13.84}$}        \\
\multicolumn{1}{c|}{\textbf{MAE}}       & $\mathbf{0.655}$     & $\mathbf {0.53}$                  & $\mathbf{0.49}$      & $\mathbf{0.69}$                        & $\mathbf{1.80}$      & $\mathbf{1.54}$      \\
\multicolumn{1}{c|}{$\mathbf{\sigma}$}  & $\mathbf{0.53}$      & $\mathbf{0.62}$                   & $\mathbf{0.74}$      & $\mathbf{1.03}$                        & $\mathbf{11.57}$     & $\mathbf{7.60}$      \\ \hline
\end{tabular}
\label{tab:error:fig8}
\end{center}
\end{table}

\begin{table}[htbp]
\caption{follow scenario: Comparison of RMSE and standard deviation between KalmanNet and IMM for each state component.}
\begin{center}
\begin{tabular}{ccccccc}
\multicolumn{1}{l|}{}                   & \multicolumn{2}{c|}{\textbf{Position}}                   & \multicolumn{2}{c|}{\textbf{Velocity}}                        & \multicolumn{2}{c}{\textbf{Acceleration}}   \\
\multicolumn{1}{c|}{\textbf{KalmanNet}} & $\mathbf{x}$         & \multicolumn{1}{c|}{$\mathbf{y}$} & $\mathbf{\dot x}$    & \multicolumn{1}{c|}{$\mathbf{\dot y}$} & $\mathbf{\ddot x}$   & $\mathbf{\ddot y}$   \\ \hline
\multicolumn{1}{c|}{\textbf{RMSE}}      & \multicolumn{2}{c}{$0.92$}                               & \multicolumn{2}{c}{$1.59$}                                    & \multicolumn{2}{c}{$12.45$}                 \\
\multicolumn{1}{c|}{\textbf{MAE}}       & $0.61$               & $\mathbf{0.50}$                            & $0.38$               & $0.49$                                 & $5.19$               & $7.58$               \\
\multicolumn{1}{c|}{$\mathbf{\sigma}$}  & $0.42$               & $0.48$                            & $1.24$               & $0.93$                                 & $6.80$               & $7.16$               \\ \hline
\multicolumn{1}{l}{}                    & \multicolumn{1}{l}{} & \multicolumn{1}{l}{}              & \multicolumn{1}{l}{} & \multicolumn{1}{l}{}                   & \multicolumn{1}{l}{} & \multicolumn{1}{l}{} \\
\multicolumn{1}{c|}{\textbf{IMM}}       & $\mathbf{x}$         & \multicolumn{1}{c|}{$\mathbf{y}$} & $\mathbf{\dot x}$    & \multicolumn{1}{c|}{$\mathbf{\dot y}$} & $\mathbf{\ddot x}$   & $\mathbf{\ddot y}$   \\ \hline
\multicolumn{1}{c|}{\textbf{RMSE}}      & \multicolumn{2}{c}{$\mathbf{0.90}$}                      & \multicolumn{2}{c}{$\mathbf{0.45}$}                           & \multicolumn{2}{c}{$\mathbf{4.19}$}         \\
\multicolumn{1}{c|}{\textbf{MAE}}       & $\mathbf{0.58}$      & $0.51$                  & $\mathbf{0.21}$      & $\mathbf{0.22}$                        & $\mathbf{0.49}$      & $\mathbf{0.47}$      \\
\multicolumn{1}{c|}{$\mathbf{\sigma}$}  & $\mathbf{0.28}$      & $\mathbf{0.51}$                   & $\mathbf{0.30}$      & $\mathbf{0.33}$                        & $\mathbf{3.53}$      & $\mathbf{2.25}$      \\ \hline
\end{tabular}
\label{tab:error:follow}
\end{center}
\end{table}

\section{Discussion}
In this work, KalmanNet\cite{revach2022kalmannet} was evaluated on real world automotive radar data. After evaluating the performance on the RadarScenes\cite{radar_scenes_dataset} dataset a detailed evaluation on two out-of distribution scenarios, the 8-drive and the follow-drive, was conducted and the results compared to a reference IMM filter. These two scenarios were selected because they represent a stress test for automotive filters with highly non-linear motion as well as a common driving scenario. The error across the entire RadarScenes dataset was measured using \textit{MAE}, its standard deviation, and \textit{RMSE}. For the two specific driving scenarios a detailed comparison to an IMM filter for the aforementioned metrics, as well as the \textit{NEES}, \textit{NIS}, and the state uncertainty matrix is provided. 

Tab. \ref{tab:error:rs} reports the errors of KalmanNet on the RadarScenes dataset, with a \textit{RMSE} of $1.13$ for position, \textit{MAE} of $0.66$ and $0.65$ in x and y, and a \textit{MAE} deviation of $0.83$ and $0.81$. These errors are within the margins given by the extent of the tracked objects and demonstrate that KalmanNet can provide a stable estimate in terms of \textit{RMSE} and \textit{MAE} of the position. However, these metrics do not provide a sufficiently conclusive evaluation of KalmanNet. In the remainder of this section the performance on the two detailed scenarios is discussed, providing further insight into the performance of KalmanNet.  

\subsection{The $8$-drive scenario}

Tab. \ref{tab:error:fig8} clearly shows that the IMM filter has a lower error and error deviations for all state components in the $8$-drive scenario. This can also be observed in Fig. \ref{fig:comp:state:8drive}. KalmanNet's \textit{RMSE} values are $1.23$, $2.98$, and $17.96$ for position, velocity, and acceleration, while the IMM filter has a \textit{RMSE} of $1.08$, $1.28$, and $13.84$, respectively. While the KalmanNet error is slightly larger for position and acceleration, it is twice as high as the error of the IMM for the velocity. In terms of the \textit{RMSE} and \textit{MAE}, the IMM filter outperforms KalmanNet, with a particularly large performance gap for the velocity estimate. 

To evaluate the statistical consistency, the \textit{NEES} and \textit{NIS} scores are shown in Fig. \ref{fig:comp:norm:8drive}. Neither KalmanNet nor the IMM filter have an estimation error that is consistent with the estimated state covariance, both have some \textit{NEES} values inside the consistency confidence interval, but they are generally above the upper interval limit, indicating that both systems underestimate their estimation uncertainty. It is further apparent, that the KalmanNet \textit{NEES} is less stable and fluctuates more than the IMM \textit{NEES} score. The \textit{NIS} score remains within the confidence interval for the IMM filter, indicating consistency between the measurement residual and the innovation covariance. The KalmanNet \textit{NIS} score, however, is above the upper critical value and subject to strong fluctuations. This reveals an inconsistency between the innovation and the innovation covariance of KalmanNet. The results of the NEES score are supported by the state uncertainty shown in Fig. \ref{fig:comp:pvol:8drive}, the IMM filter has a small state uncertainty and while the state uncertainty of KalmanNet is slightly higher, it fluctuates strongly.

\subsection{The follow-drive scenario}
As reported above, both KalmanNet and IMM perform better on the follow-drive scenario than on the 8-drive scenario. However, the IMM filter consistently shows lower errors and error deviations than KalmanNet across all state components. Fig. \ref{fig:comp:state:foll_drive} and Tab. \ref{tab:error:follow} show that KalmanNet has a \textit{RMSE} of $0.92$, $1.59$, and $12.45$, while the IMM filter has a \textit{RMSE} of $0.90$, $0.45$, and $4.19$, for position, velocity, and acceleration. In terms of mean position error, the filters have a similar performance, although KalmanNet is again characterized by a highly unstable error. Analogous to the $8$-drive scenario, the performance difference between KalmanNet and the IMM filter increases dramatically for velocity and acceleration, where the IMM \textit{RMSE} is smaller than the KalmanNet \textit{RMSE} by a factor of three. 

The results discussed above demonstrate that KalmanNet has lower accuracy and lower precision than a linear IMM filter, both for the follow-drive and for the non-linear motion of the $8$-drive scenario. These findings contradict the claim made by \cite{revach2022kalmannet} that KalmanNet, using an approximated motion model and unknown noise statistics, outperforms classical MB KF algorithms by learning the noise distributions from a limited amount of data. It is important to emphasize that the IMM filter was not fine-tuned for these scenarios, suggesting that its performance could be further enhanced with additional optimization. In contrast, such potential for significant improvement is limited in the case of KalmanNet. While increasing the amount of training data or further modifications to the loss function may yield marginal improvements, they are unlikely to enable KalmanNet to effectively learn the underlying noise distributions. 

Furthermore, the state uncertainty matrix and \textit{NEES} metrics indicate an unstable state covariance estimate and inconsistency between the estimated error covariance and the actual error covariance. Finally, the inconsistent \textit{NIS} score suggests that KalmanNet failed to properly learn the motion model and noise statistics from the available data. This suggests that KalmanNet lacks the reliability and robustness needed for safety-critical systems, such as ADAS. However, these results also show that KalmanNet can provide a method for filtering that does not require full knowledge of the dynamical system or expert tuning of the filter. This could be advantageous in less critical applications.

\section{Conclusion}
In this work we presented a comprehensive evaluation of KalmanNet\cite{revach2022kalmannet} on automotive radar data and compared it to an IMM filter. We have trained KalmanNet using the RadarScenes\cite{radar_scenes_dataset} dataset and evaluated it using the labeled radar points from RadarScenes as well as out of distribution data with precise ground truth measurements of the OV. Our evaluation is in terms of \textit{RMSE}, \textit{MAE}, the normalized errors \textit{NEES}, \textit{NIS}, and the state uncertainty. Our results showed that KalmanNet had a lower accuracy, less stable tracking results, and a higher uncertainty than the reference IMM filter. While KalmanNet might be promising for certain applications and benefits from the independence of defined SS models, our findings indicate an unfitness for safety critical applications such as ADAS systems. 

\bibliographystyle{unsrt}
\bibliography{literature}

\vspace{12pt}

\end{document}